\documentclass[conference]{IEEEtran}
\IEEEoverridecommandlockouts

\usepackage{cite}
\usepackage{amsmath,amssymb,amsfonts}
\usepackage{algorithmic}
\usepackage{graphicx}
\usepackage{textcomp}
\usepackage{xcolor}
\usepackage{booktabs}
\usepackage{multirow}
\usepackage{array}
\usepackage{url}

\def\BibTeX{{\rm B\kern-.05em{\sc i\kern-.025em b}\kern-.08em
    T\kern-.1667em\lower.7ex\hbox{E}\kern-.125emX}}

\begin{document}

\title{Multi-Modal Opinion Integration for Financial Sentiment Analysis using Cross-Modal Attention}

\author{\IEEEauthorblockN{1\textsuperscript{st} Yujing Liu}
\IEEEauthorblockA{\textit{College of Computing} \\
\textit{Georgia Institute of Technology}\\
Atlanta, USA\\
yliu3393@gatech.edu}
\and
\IEEEauthorblockN{2\textsuperscript{nd} Chen Yang}
\IEEEauthorblockA{\textit{College of Engineering} \\
\textit{University of Pennsylvania} \\
Philadelphia, USA \\
sophiacy@upenn.edu}
}
\maketitle

\begin{abstract}
In recent years, financial sentiment analysis of public opinion has become increasingly important for market forecasting and risk assessment. However, existing methods often struggle to effectively integrate diverse opinion modalities and capture fine-grained interactions across them. This paper proposes an end-to-end deep learning framework that integrates two distinct modalities of financial opinions—recency modality (timely opinions) and popularity modality (trending opinions)—through a novel cross-modal attention mechanism specifically designed for financial sentiment analysis. While both modalities consist of textual data, they represent fundamentally different information channels: recency-driven market updates versus popularity-driven collective sentiment. Our model first uses BERT (Chinese-wwm-ext) for feature embedding and then employs our proposed Financial Multi-Head Cross-Attention (FMHCA) structure to facilitate information exchange between these distinct opinion modalities. The processed features are optimized through a transformer layer and fused using multimodal factored bilinear pooling for classification into negative, neutral, and positive sentiment. Extensive experiments on a comprehensive dataset covering 837 companies demonstrate that our approach achieves an accuracy of 83.5\%, significantly outperforming baselines including BERT+Transformer by 21\%. These results highlight the potential of our framework to support more accurate financial decision-making and risk management.
\end{abstract}


\section{Introduction}

Financial sentiment analysis has become an important component of quantitative finance, enabling automated assessment of market sentiment from massive text data. Traditional financial sentiment analysis methods rely on single sources of analysis, such as news articles or social media posts exclusively. Such single-modality analysis often neglects the complex interactions between different types of opinion sources and fails to capture the synergistic impact of diverse information channels on market dynamics.

The digitalization of financial information presents unique opportunities for multi-modal analysis. We identify two distinct modalities of financial opinions that serve different functions in market sentiment formation: (1) \emph{recency modality} (timely opinions) from news articles, analyst reports, and official announcements, which provide timely market insights and factual updates, and (2) \emph{popularity modality} (trending opinions) from viral social media discussions and popular forum content, which capture the collective market psychology and consensus sentiment. Although both modalities consist of textual data, they represent fundamentally different information channels in financial discourse—immediate factual updates versus aggregated collective sentiment. Understanding the cross-modal relationships between these modalities is crucial for comprehensive sentiment analysis. 

Deep learning has made numerous advancements in recent years. Transformer-based models, such as BERT, have achieved remarkable success in natural language processing tasks. However, directly applying these models to financial sentiment analysis still faces several limitations: (1) It is difficult to effectively model the cross-modal relationships between different types of viewpoints; (2) The fusion mechanism for heterogeneous text features is not effective; and (3) There is a lack of domain-specific adaptability for financial contexts. 

To address these challenges, we propose a novel architecture that combines BERT's representation capabilities with our specifically designed Financial Multi-Head Cross-Attention (FMHCA) mechanism and multimodal fusion techniques. Our approach is the first to explicitly model the cross-modal relationships between recency and popularity modalities in financial sentiment analysis. Our key contributions are:

\begin{itemize}
\item A novel multi-modal framework that recognizes and leverages two distinct modalities in financial opinions: recency modality (timely opinions) and popularity modality (trending opinions), representing different information channels despite both being textual
\item An end-to-end deep learning architecture that integrates BERT embeddings, our proposed FMHCA, transformer processing, and multimodal factorized bilinear pooling for enhanced financial sentiment classification
\item The Financial Multi-Head Cross-Attention (FMHCA) mechanism, specifically designed to capture asymmetric cross-modal relationships between recency-driven factual updates and popularity-driven collective sentiment in financial contexts
\item Comprehensive experimental validation on a large-scale dataset of 837 companies, achieving state-of-the-art performance with 83.5\% accuracy and demonstrating significant improvements over single-modality baselines
\end{itemize}


\section{Related Work}

Financial sentiment analysis has seen significant advances in recent years, driven by the development of transformer-based language models and the increasing availability of multimodal data. Early work primarily focused on textual data, where transformer models such as BERT and FinBERT have achieved strong performance in financial sentiment classification \cite{araci2019finbert,hajek2022predicting,gossi2023fintuned}. More recent studies fine-tune pretrained models specifically for financial applications, demonstrating improvements in sentiment classification and volatility prediction tasks \cite{niu-etal-2023-kefvp,feng-etal-2024-knowledge}.

Beyond text, multimodal sentiment analysis has emerged as an important research direction. Models that integrate information across textual, visual, and acoustic modalities have demonstrated that cross-modal interactions significantly improve performance compared to single-modality baselines \cite{zhu2023multimodal,quan2022mghf,guo2024cmdaf}. Cross-attention mechanisms in particular have been widely adopted to enhance inter-modal information exchange, enabling models to better capture complementary cues across modalities \cite{wang2024aspect,feng-etal-2024-knowledge}.

In the financial domain, researchers have also begun incorporating temporal dynamics and opinion diversity into sentiment models. For instance, Cao et al. \cite{cao2023financial} proposed capturing both ``timely'' and ``trending'' opinions from microblogs to improve stock-related sentiment prediction, showing that public opinion dynamics provide complementary signals beyond static features. Other works introduce fine-grained emotion datasets such as StockEmotions, which integrates investor emotions with multivariate time series \cite{lee2023stockemotions}, and SEntFiN, which provides entity-aware annotations for financial sentiment \cite{sinha2023sentfin}. Large-scale datasets such as FNSPID \cite{dong2024fnspid} further enable benchmarking across long time horizons and diverse companies.

Another line of work addresses practical challenges of applying sentiment models in real-world finance. For example, studies on earnings call transcripts explore sparse attention mechanisms and augmentation strategies to handle long, noisy documents with limited labeled data \cite{yuan2023earnings}. Graph-based models such as DialogueGAT capture conversational structures in conference calls to support financial risk prediction \cite{sang-bao-2022-dialoguegat}. Similarly, FinBERT variants have been used to support predictive tasks including exchange rate forecasting \cite{hajek2022predicting} and fine-tuned sentiment modeling \cite{gossi2023fintuned}.

Despite these advances, existing methods still face limitations. Most multimodal sentiment models are developed for general-purpose datasets and do not adequately address the integration of diverse opinion modalities specific to finance. Few approaches explicitly capture both fresh and popular opinion streams while enabling fine-grained cross-modal interactions. Furthermore, large-scale evaluations across companies remain scarce, and issues such as robustness, class imbalance, and deployment constraints in financial institutions require further exploration.

\section{Methodology}

\begin{figure}[htbp]
\centerline{\includegraphics[scale=0.15]{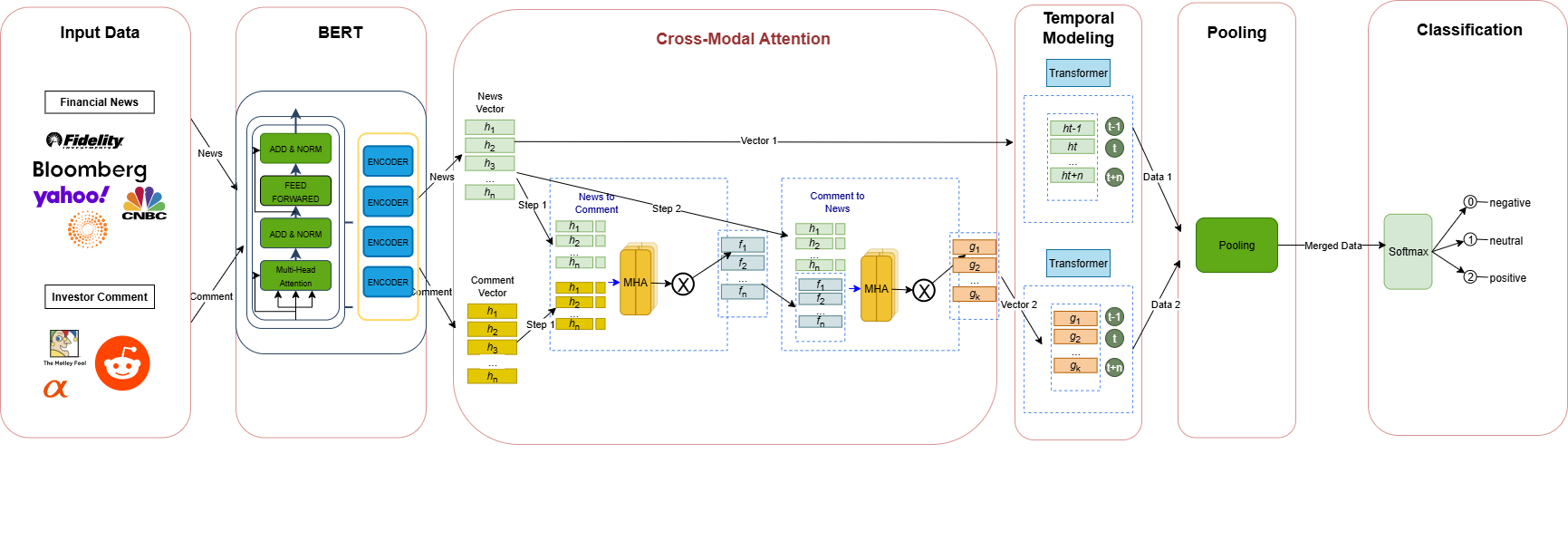}}
\caption{Overview of the proposed cross-modal attention enhanced financial sentiment analysis architecture. The model processes timely and trending opinions through BERT embedding, applies cross-modal attention mechanisms (MHA), processes through transformer layers, and fuses representations using MFB before final classification.}
\label{fig:architecture}
\end{figure}

The architecture we propose consists of multiple components for effective cross-modal sentiment analysis. Figure \ref{fig:architecture} shows the overall design of the system, which processes fresh and popular viewpoints through parallel paths before fusing these representations for final classification.

\subsection{Problem Formulation}

We focus on financial sentiment analysis across a comprehensive dataset containing $N$ publicly listed companies. For each company $c_i$ where $i \in \{1,2,\ldots,N\}$, we define and collect two distinct modalities of financial opinions from social media and news platforms:

\textbf{Modality 1 - Recency Modality (timely opinions):} This modality captures newly released information such as news articles, analyst reports, earnings announcements, and regulatory filings. These opinions reflect the timeliness aspect of market information and represent immediate market reactions to recent developments.

\textbf{Modality 2 - Popularity Modality (Trending Opinions):} This modality captures widely discussed content or topics on social media platforms and financial forums during a specific time period. These opinions reflect the collective market sentiment and represent the viral or consensus aspects of market discourse.

Although both modalities consist of textual data, they represent fundamentally different information channels in financial markets: recency-based information flow versus popularity-based collective sentiment. The number of opinions in each modality varies significantly across companies due to heterogeneous market attention and discussion intensity, with some companies having extensive coverage while others receive limited attention.

Formally, let $\mathbf{F}_i \in \mathbb{R}^{m_i \times d}$ denote the feature matrix of timely opinions for company $c_i$, where $m_i$ represents the number of timely opinions associated with company $c_i$ and $d$ is the embedding dimension. Each row $\mathbf{F}_i[j,:]$ corresponds to the $d$-dimensional feature representation of the $j$-th timely opinion. Similarly, let $\mathbf{H}_i \in \mathbb{R}^{n_i \times d}$ represent the feature matrix of trending opinions for company $c_i$, where $n_i$ denotes the number of trending opinions. Each row $\mathbf{H}_i[k,:]$ represents the $d$-dimensional feature vector of the $k$-th trending opinion.

The dimensions $m_i$ and $n_i$ vary across companies, reflecting the heterogeneous nature of market discourse:
\begin{align}
m_i &\in \{1, 2, \ldots, M_{\max}\} \quad \text{(timely opinion count for company $i$)} \\
n_i &\in \{1, 2, \ldots, N_{\max}\} \quad \text{(trending opinion count for company $i$)}
\end{align}
where $M_{\max}$ and $N_{\max}$ represent the maximum number of timely and trending opinions observed across all companies in the dataset.

Thus, each company $c_i$ is represented by a heterogeneous opinion pair $(\mathbf{F}_i, \mathbf{H}_i)$ capturing both timely and trending market perspectives. The objective of financial sentiment analysis is to learn a mapping function:
\begin{equation}
\mathcal{M}: (\mathbf{F}_i, \mathbf{H}_i) \mapsto y_i
\end{equation}
where $y_i \in \{-1,0,+1\}$ corresponds to negative, neutral, and positive sentiment labels, respectively.

Multi-Modal Nature of Text-Based Financial Opinions: While both $\mathbf{F}_i$ and $\mathbf{H}_i$ consist of textual data, they constitute distinct modalities in the context of financial sentiment analysis due to their fundamentally different characteristics: (1) Temporal characteristics. timely opinions have immediate temporal relevance, while trending opinions aggregate sentiment over time periods. (2) Source diversity. timely opinions originate from authoritative sources (news agencies, analysts), while trending opinions emerge from collective user-generated content. (3) Information function. timely opinions convey factual market updates, while trending opinions reflect collective market psychology and consensus. (4) Content structure. timely opinions follow journalistic or analytical formats, while trending opinions exhibit conversational and viral content patterns. (5) Market impact. timely opinions drive immediate market reactions, while trending opinions capture sustained market sentiment.

The central challenge lies in effectively integrating these heterogeneous opinion modalities with varying dimensions $(m_i \times d)$ and $(n_i \times d)$, while preserving and leveraging the complementary information embedded in both $\mathbf{F}_i$ and $\mathbf{H}_i$ matrices through cross-modal learning.

\subsection{BERT Feature Extraction}

We use BERT (Chinese-wwm-ext) as the foundation for extracting contextual representations from both timely and trending opinions. For each company $c_i$ with opinion matrices $\mathbf{F}_i$ and $\mathbf{H}_i$, BERT generates contextualized embeddings as follows:

\begin{equation}
\mathbf{E}_f^{(i)} = \text{BERT}(\mathbf{F}_i), \quad \mathbf{E}_h^{(i)} = \text{BERT}(\mathbf{H}_i)
\end{equation}

where $\mathbf{E}_f^{(i)} \in \mathbb{R}^{m_i \times d_{bert}}$ and $\mathbf{E}_h^{(i)} \in \mathbb{R}^{n_i \times d_{bert}}$ represent the contextualized embeddings for timely and trending opinions respectively, with $d_{bert} = 768$ being the BERT hidden dimension.

The [CLS] tokens from each modality are extracted and projected to a lower-dimensional space:

\begin{equation}
\mathbf{f}_i = \text{Project}(\mathbf{E}_f^{(i)}[0,:]), \quad \mathbf{h}_i = \text{Project}(\mathbf{E}_h^{(i)}[0,:])
\end{equation}

where the projection layer maps from $\mathbb{R}^{768}$ to $\mathbb{R}^{128}$ with ReLU activation, and $\mathbf{f}_i, \mathbf{h}_i \in \mathbb{R}^{128}$ represent the projected representations for company $c_i$.

\subsection{Financial Multi-Head Cross-Attention (FMHCA)}

The core innovation and primary contribution of our approach is the Financial Multi-Head Cross-Attention (FMHCA) mechanism, which we propose specifically for multi-modal financial sentiment analysis. Unlike existing attention mechanisms that operate within single modalities, our FMHCA is designed to capture and leverage the complementary relationships between two fundamentally different information modalities in financial markets: recency-driven timely opinions and popularity-driven trending opinions.

The key novelty of FMHCA lies in its ability to model the asymmetric information flow between these modalities, recognizing that timely opinions provide factual market updates while trending opinions capture collective market psychology. This cross-modal interaction is crucial in financial contexts where market sentiment emerges from the interplay between new information (recency modality) and collective market reactions (popularity modality). The FMHCA operates through a novel two-stage attention process specifically designed to exploit this complementary nature.

\textbf{Stage 1: Trending-to-Timely Attention}
The first attention stage allows trending opinions to query relevant information from timely opinions:

\begin{equation}
\mathbf{s}_1^{(i)}, \mathbf{G}^{(i)} = \text{FMHCA}(\mathbf{H}_i, \mathbf{F}_i)
\end{equation}

where $\mathbf{H}_i \in \mathbb{R}^{n_i \times d}$ and $\mathbf{F}_i \in \mathbb{R}^{m_i \times d}$ are the embedded representations with positional encoding and CLS tokens for company $c_i$, $\mathbf{s}_1^{(i)} \in \mathbb{R}^{n_i \times m_i}$ represents the attention weights, and $\mathbf{G}^{(i)} \in \mathbb{R}^{n_i \times d}$ is the attended representation.

\textbf{Stage 2: Attended-to-Trending Attention}
The second stage performs attention from the first stage output back to the trending opinions:

\begin{equation}
\mathbf{s}_2^{(i)}, \mathbf{F}_i' = \text{FMHCA}(\mathbf{G}^{(i)}, \mathbf{H}_i)
\end{equation}

where $\mathbf{F}_i' \in \mathbb{R}^{n_i \times d}$ represents the refined timely opinion representation after cross-modal interaction.

This novel two-stage cross-modal attention process represents a significant advancement over traditional single-modality approaches by: (1) enabling bidirectional information flow between recency and popularity modalities, (2) preserving the semantic coherence of each modality while facilitating cross-modal learning, and (3) capturing the temporal dynamics inherent in financial sentiment formation where timely information influences trending discussions and vice versa.

The Financial Multi-Head Cross-Attention mechanism is defined as:

\begin{equation}
\text{FMHCA}(\mathbf{Q}, \mathbf{K}, \mathbf{V}) = \text{Concat}(\text{head}_1, \ldots, \text{head}_h)\mathbf{W}^O
\end{equation}

where each attention head is computed as:

\begin{equation}
\text{head}_j = \text{Attention}(\mathbf{Q}\mathbf{W}_j^Q, \mathbf{K}\mathbf{W}_j^K, \mathbf{V}\mathbf{W}_j^V)
\end{equation}

\begin{equation}
\text{Attention}(\mathbf{Q}, \mathbf{K}, \mathbf{V}) = \text{softmax}\left(\frac{\mathbf{Q}\mathbf{K}^T}{\sqrt{d_k}}\right)\mathbf{V}
\end{equation}

where $h$ is the number of attention heads, $\mathbf{W}_j^Q, \mathbf{W}_j^K, \mathbf{W}_j^V \in \mathbb{R}^{d \times d_k}$ are learnable parameter matrices for the $j$-th head, $\mathbf{W}^O \in \mathbb{R}^{hd_k \times d}$ is the output projection matrix, and $d_k = d/h$ is the dimension of each attention head.

\subsection{Transformer Processing}

Following the FMHCA stages, both the attended representation $\mathbf{F}_i'$ and the original trending representation $\mathbf{H}_i$ are processed through individual transformer layers to refine the features and capture temporal dependencies:

\begin{equation}
\mathbf{F}_i'' = \text{Transformer}(\mathbf{F}_i'), \quad \mathbf{H}_i' = \text{Transformer}(\mathbf{H}_i)
\end{equation}

where $\mathbf{F}_i'' \in \mathbb{R}^{n_i \times d}$ and $\mathbf{H}_i' \in \mathbb{R}^{n_i \times d}$ represent the refined representations for company $c_i$. Each transformer layer consists of multi-head self-attention and position-wise feed-forward networks with residual connections and layer normalization:

\begin{align}
\mathbf{X}' &= \text{LayerNorm}(\mathbf{X} + \text{MHSA}(\mathbf{X})) \\
\text{Transformer}(\mathbf{X}) &= \text{LayerNorm}(\mathbf{X}' + \text{FFN}(\mathbf{X}'))
\end{align}

where $\text{MHSA}$ denotes multi-head self-attention and $\text{FFN}$ represents the feed-forward network.

\subsection{Multimodal Factorized Bilinear Pooling}

To effectively fuse the processed representations, we employ Multimodal Factorized Bilinear Pooling (MFB), which captures complex interactions between modalities while maintaining computational efficiency:

\begin{equation}
\mathbf{z}_i = \text{MFB}([\mathbf{F}_{i,\text{cls}}'', \mathbf{H}_{i,\text{cls}}'])
\end{equation}

where $\mathbf{F}_{i,\text{cls}}'' \in \mathbb{R}^d$ and $\mathbf{H}_{i,\text{cls}}' \in \mathbb{R}^d$ represent the [CLS] token representations from the transformer-processed features for company $c_i$.

The MFB operation is defined as:

\begin{equation}
\mathbf{z}_i = \sum_{k=1}^{K} (\mathbf{W}_f^{(k)}\mathbf{F}_{i,\text{cls}}'') \odot (\mathbf{W}_h^{(k)}\mathbf{H}_{i,\text{cls}}')
\end{equation}

where $\odot$ denotes element-wise multiplication, $\mathbf{W}_f^{(k)}, \mathbf{W}_h^{(k)} \in \mathbb{R}^{d_{mfb} \times d}$ are learnable projection matrices for the $k$-th factor, $K$ is the factorization parameter, and $\mathbf{z}_i \in \mathbb{R}^{d_{mfb}}$ represents the fused multimodal representation for company $c_i$.

\subsection{Classification Layer}

The fused representation is passed through a final linear layer for sentiment classification:

\begin{equation}
\mathbf{p}_i = \text{softmax}(\mathbf{W}_c\mathbf{z}_i + \mathbf{b}_c)
\end{equation}

where $\mathbf{W}_c \in \mathbb{R}^{3 \times d_{mfb}}$ and $\mathbf{b}_c \in \mathbb{R}^3$ are learnable parameters, and $\mathbf{p}_i \in \mathbb{R}^3$ represents the probability distribution over the three sentiment classes (negative, neutral, positive) for company $c_i$. The predicted sentiment label is obtained as:

\begin{equation}
\hat{y}_i = \arg\max_{j \in \{-1,0,+1\}} \mathbf{p}_i[j]
\end{equation}

where $\mathbf{p}_i[j]$ denotes the $j$-th component of the probability vector corresponding to sentiment class $j$.

\section{Experimental Results}

\subsection{Dataset}

We conduct experiments on a dataset containing financial opinions for 837 companies. The dataset includes both timely opinions (recent news articles, press releases, analyst reports) and trending opinions (trending social media posts, popular forum discussions, viral financial content). Each sample was manually annotated with sentiment labels (negative, neutral, positive).

The dataset characteristics are as follows:
\begin{itemize}
\item Number of companies: 837
\item Average opinions per company: 150-300 per modality
\item Label distribution: Negative (32\%), Neutral (41\%), Positive (27\%)
\item Time span: 2020-2023
\end{itemize}

\subsection{Implementation Details}

Our model is implemented using PyTorch and trained on NVIDIA RTX 4090 GPUs. The key hyperparameters include chinese-bert-wwm-ext as the base model, an embedding dimension of 128, 8 attention heads, and MFB factorization parameter of 16. Training is conducted with a batch size of 16, learning rate of 2e-5 using Adam optimizer, over 50 epochs with a dropout rate of 0.1.

We compare our approach against several baseline methods including standard BERT with classification head, BERT with multi-layer perceptron (BERT + MLP), BERT embeddings processed by LSTM layers (BERT + LSTM), and BERT with additional transformer layers (BERT + Transformer). We also evaluate fusion-based approaches including LSTM-based fusion of modalities (BERT + LSTM + Fusion) and transformer-based fusion without cross-modal attention (BERT + Transformer + Fusion).

Model performance is evaluated using the following metrics:
\begin{itemize}
\item Accuracy: Overall classification accuracy
\item Weighted Average F1: F1-score weighted by class support
\item Weighted Average Precision: Precision weighted by class support  
\item Weighted Average Recall: Recall weighted by class support
\end{itemize}

\subsection{Main Results}

Table~\ref{tab:main_results} presents the comparative performance of our proposed model against the baseline methods. Our model achieves superior performance across all metrics, with 83.5\% accuracy, 82.0\% weighted F1-score, 82.0\% weighted precision, and 81.0\% weighted recall.

\begin{table}[htbp]
\caption{Comparison with Baseline Methods}
\label{tab:main_results}
\begin{center}
\resizebox{\columnwidth}{!}{%
\begin{tabular}{|c|c|c|c|c|}
\hline
\textbf{Method} & \textbf{Acc.} & \textbf{F1} & \textbf{Prec.} & \textbf{Recall} \\
\hline
BERT & 62.5\% & 62.3\% & 62.3\% & 62.5\% \\
\hline
BERT + MLP & 57.5\% & 51.1\% & 52.2\% & 57.5\% \\
\hline
BERT + LSTM & 57.0\% & 48.1\% & 61.8\% & 57.0\% \\
\hline
BERT + Transformer & 62.5\% & 62.4\% & 64.9\% & 62.5\% \\
\hline
BERT + LSTM + Fusion & 47.5\% & 30.6\% & 22.6\% & 47.5\% \\
\hline
BERT + Transformer + Fusion & 77.0\% & 77.1\% & 77.2\% & 77.0\% \\
\hline
\textbf{Our Model} & \textbf{83.5\%} & \textbf{82.0\%} & \textbf{82.0\%} & \textbf{81.0\%} \\
\hline
\end{tabular}%
}
\end{center}
\end{table}

The results demonstrate significant improvements over all baseline methods. In particular, our model outperforms the best baseline (BERT + Transformer + Fusion) by 6.5 percentage points in accuracy, highlighting the effectiveness of the cross-modal attention mechanism.

\subsection{Ablation Study}

To understand the contribution of individual components, we conducted ablation studies by removing key architectural elements. Table~\ref{tab:ablation} shows the results.

\begin{table}[htbp]
\caption{Ablation Study Results}
\begin{center}
\label{tab:ablation}
\resizebox{\columnwidth}{!}{%
\begin{tabular}{|c|c|c|c|c|}
\hline
\textbf{Model Variant} & \textbf{Accuracy} & \textbf{W. Avg F1} & \textbf{W. Avg Prec.} & \textbf{W. Avg Recall} \\
\hline
w/o Cross-Modal Attention (MHA) & 77.0\% & 77.2\% & 77.0\% & 77.1\% \\
\hline
w/o Fusion Layer & 76.5\% & 71.9\% & 71.4\% & 76.5\% \\
\hline
\textbf{Full Model} & \textbf{83.5\%} & \textbf{82.0\%} & \textbf{82.0\%} & \textbf{81.0\%} \\
\hline
\end{tabular}
}
\end{center}
\end{table}

The ablation study reveals: (1) Removing cross-modal attention (MHA) reduces accuracy by 6.5\%, demonstrating its crucial role in modeling inter-modality relationships. (2) Removing the fusion layer decreases accuracy by 7.0\%, demonstrating the importance of effective multimodal integration. Both components contribute significantly to the model's overall performance.

\subsection{Model Robustness Analysis}

To validate that performance improvements are not solely attributed to the specific BERT variant, we evaluate our model using different Chinese BERT models. Table~\ref{tab:model_variants} shows the results.

\begin{table}[htbp]
\begin{center}
\caption{Performance Across Different BERT Models}
\resizebox{\columnwidth}{!}{%
\begin{tabular}{|c|c|c|c|c|}
\hline
\textbf{BERT Model} & \textbf{Accuracy} & \textbf{W. Avg F1} & \textbf{W. Avg Prec.} & \textbf{W. Avg Recall} \\
\hline
Chinese & 83.0\% & 81.0\% & 81.0\% & 79.0\% \\
\hline
Chinese-wwm & 82.5\% & 80.0\% & 82.0\% & 81.0\% \\
\hline
Chinese-wwm-ext & \textbf{83.5\%} & \textbf{82.0\%} & \textbf{82.0\%} & \textbf{81.0\%} \\
\hline
\end{tabular}
}
\label{tab:model_variants}
\end{center}
\end{table}

The consistent performance across different BERT variants (with only 1\% variance) proves that our model's improvements result from the architectural innovations rather than the choice of pre-trained language model.

\subsection{Performance Analysis}

\subsubsection{Cross-Modal Attention Effectiveness}

The cross-modal attention mechanism enables the model to identify and leverage complementary information between timely and trending opinions. Analysis of attention weights indicates that the model learns to focus on: (1) Temporal relevance: timely opinions provide recent market developments. (2) Sentiment intensity: trending opinions capture emotional market reactions. (3) Contextual coherence: Cross-modal attention aligns related concepts across modalities.

\subsubsection{Fusion Strategy Impact}

The MFB fusion mechanism proves more effective than simple concatenation or addition approaches. The factorized bilinear pooling captures complex interactions while maintaining computational efficiency, leading to a 7\% improvement in accuracy compared to the model without fusion.

These findings suggest directions for future improvements, including specialized training for financial sarcasm detection and temporal sentiment modeling.

\section{Discussion}

Our research integrates multiple perspectives for financial sentiment analysis. The cross-modal attention mechanism can utilize timely timely and trending opinions to gain a more detailed understanding of the market sentiment. The excellent performance of this model demonstrates its significant practical application value in real-time market sentiment monitoring, risk assessment, investment decision support, and regulatory compliance systems.

From a methodological perspective, our work provides a novel approach for the application of cross-modal attention in financial text analysis, while demonstrating effective multimodal fusion techniques for heterogeneous text sources. At the same time, we conducted an in-depth analysis of the relative importance of different components in financial sentiment analysis tasks through a comprehensive evaluation framework.

However, although we tried to make this dataset as comprehensive as possible, it is still limited to the Chinese market and may not be applicable to other language environments. The model currently does not take into account the credibility of the source and ignores the time series effects present in the opinions. Future research should explore multilingual extensions, source credibility weighting, temporal emotion modeling, and real-time deployment capabilities.

\section{Conclusion}

This article proposes a deep learning architecture for financial sentiment analysis, which effectively integrates fresh and popular public opinion through cross-modal attention mechanisms. Our method combines BERT based feature extraction with multi-head cross-attention mechanism and multimodal factorization bilinear pooling, achieving an accuracy of 83.5\% on a comprehensive dataset containing 837 companies, significantly better than the baseline method.

The cross-modal attention mechanism improves accuracy by 6.5\% through effective information exchange between opinion modalities, while the fusion mechanism achieves an additional 7\% improvement through complex multimodal integration. The model performance remains robust across different BERT variants, demonstrating the effectiveness of our architectural innovation rather than relying on specific pre trained models.

Its practical significance goes beyond academic research and lays the foundation for advanced financial sentiment analysis systems that support investment decision-making, risk management, and market monitoring. As financial markets increasingly rely on automated sentiment analysis, our work represents an important step towards more complex integration of public opinion, opening up new avenues for multimodal analysis in financial applications.

\bibliographystyle{ieeetr}   
\bibliography{references}    
\end{document}